\documentclass[10pt,twocolumn,letterpaper]{article}

\usepackage{cvpr}
\usepackage{times}
\usepackage{epsfig}
\usepackage{graphicx}
\usepackage{amsmath}
\usepackage{amssymb}

\usepackage{grffile}
\usepackage{multirow}

\usepackage{tabularx}

\newcommand{\real}{\mathbb{R}}
\newcommand{\bx}{\mathbf{x}}
\newcommand{\by}{\mathbf{y}}

\newcommand{\bff}{\mathbf{f}}

\newcommand{\bone}{\mathbf{1}}

\usepackage{color}

 \usepackage{booktabs} 
\usepackage{setspace}
\usepackage{subfigure}
\usepackage{paralist}
\usepackage[numbers,sort]{natbib} 
 
\usepackage[pagebackref=true,breaklinks=true,letterpaper=true,colorlinks,bookmarks=false]{hyperref}

 \cvprfinalcopy 

\def\cvprPaperID{1168} 

\ifcvprfinal\fi

\author{Christoph Feichtenhofer\\Graz University of Technology\\ {\tt \small \href{mailto:feichtenhofer@tugraz.at}{\textcolor{black}{feichtenhofer@tugraz.at}}} \and	Axel Pinz\\Graz University of Technology\\{\tt \small \href{mailto:axel.pinz@tugraz.at}{\textcolor{black}{axel.pinz@tugraz.at}}} \and Andrew Zisserman\\
	University of Oxford\\{\tt \small  \href{mailto:az@robots.ox.ac.uk}{\textcolor{black}{az@robots.ox.ac.uk}}}
	}
\begin{document}

\title{Convolutional Two-Stream Network Fusion for Video Action Recognition}

\maketitle

\begin{abstract}
	
Recent applications of Convolutional Neural
Networks (ConvNets) for human action recognition in videos
 have proposed different solutions for
incorporating the appearance and motion information.
We study a number of ways of fusing ConvNet towers both spatially and
temporally in order to best
take advantage of this spatio-temporal information. We make the following
findings: (i) that rather than fusing at the softmax layer,
a spatial and temporal network can be fused
at a convolution layer without loss of performance, but with a substantial
saving in parameters; (ii) that it is better to fuse such networks spatially at 
the last convolutional layer than earlier, and that additionally fusing at the class prediction layer can boost accuracy; finally (iii) that pooling of abstract convolutional features over spatiotemporal neighbourhoods further boosts performance.
Based on these studies we propose a new ConvNet architecture for spatiotemporal fusion of video snippets, and evaluate its performance on standard benchmarks where this architecture achieves state-of-the-art results.
Our code and models are available at \href{http://www.robots.ox.ac.uk/~vgg/software/two_stream_action/}{ http://www.robots.ox.ac.uk/~vgg/software/two\_stream\_action}

\vspace{-10pt}
\end{abstract}

  	\section{Introduction}
\label{sec:intro}

Action recognition in video is a highly active area of research with
state of the art systems still being far from human performance. 
As with other areas of computer vision, recent work has concentrated
on applying Convolutional Neural
Networks (ConvNets) to this task, with progress over a number of
strands: learning local spatiotemporal 
filters~\cite{Karpathy14,Taylor10,C3DICCV2015}),
incorporating optical flow
snippets~\cite{Simonyan14b}, and modelling more extended temporal 
sequences~\cite{donahue2015long,ng2015beyond}. 

However, action recognition has not yet seen the substantial gains in
performance that have been achieved in other areas by ConvNets, e.g.\
image classification~\cite{Krizhevsky12,simonyan2014very,szegedy2014going}, 
human face recognition~\cite{Schroff15},
and human pose estimation~\cite{tompson2014efficient}. Indeed
the current state of the art performance~\cite{C3DICCV2015,wang2015action} on
standard benchmarks such as UCF-101~\cite{UCF101} and
HMDB51~\cite{kuehne2011hmdb} is achieved by a combination of ConvNets
and a Fisher Vector
encoding~\cite{Perronnin10a} of 
hand-crafted features (such as HOF~\cite{Laptev08} over dense
trajectories~\cite{wangICCV13}).

\begin{figure}[!t]
	\centering
	\resizebox {0.45\textwidth }{!}{ 
		\includegraphics[width=1\textwidth]	{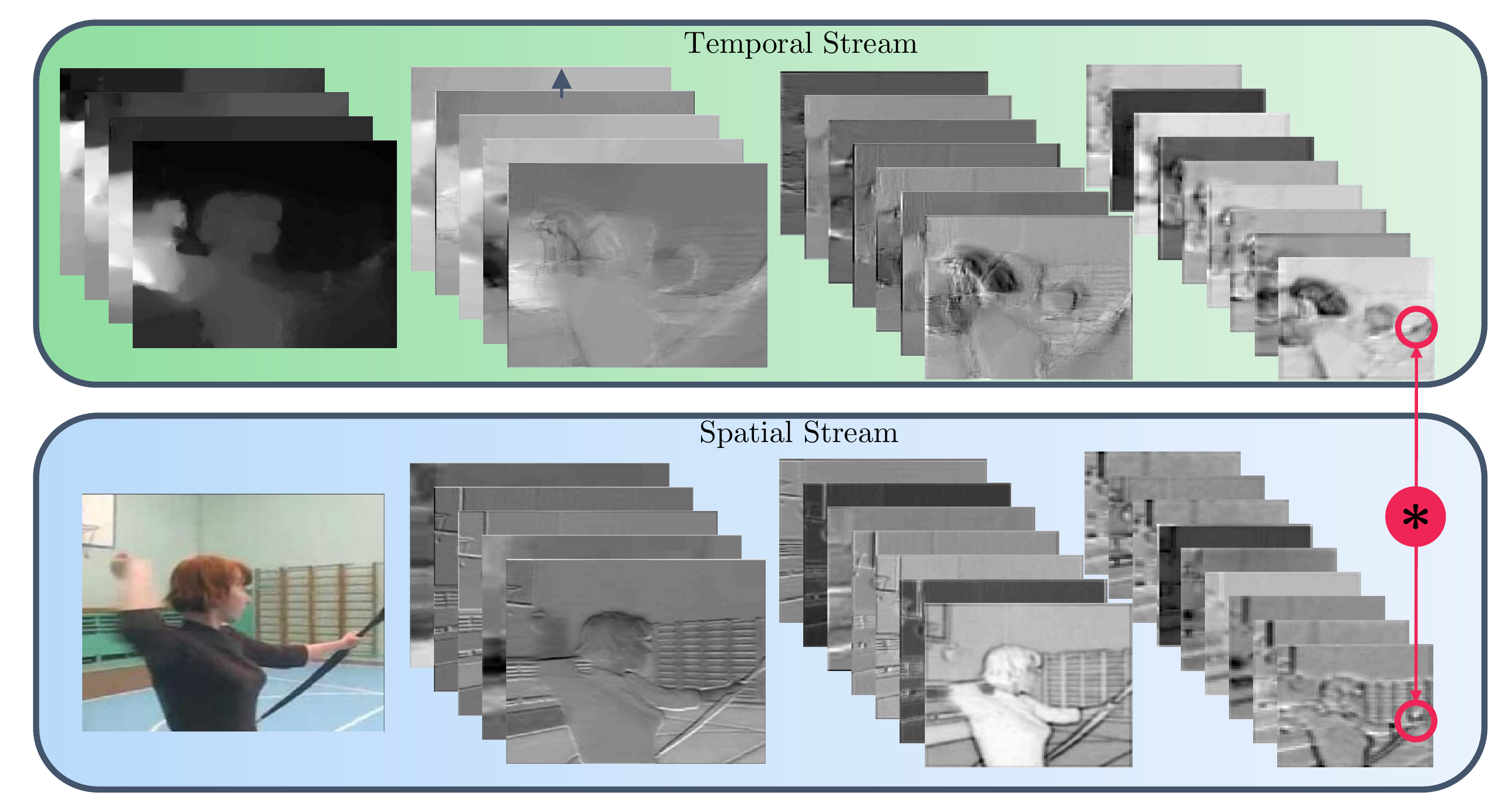}
	}
	\vspace{-5pt}
	\caption{Example outputs of the first three convolutional layers from a two-stream ConvNet model \cite{Simonyan14b}. The two networks separately capture spatial (appearance) and temporal information at a fine temporal scale. In this work we investigate several approaches to fuse the two networks over space and time. 
	}
	\vspace{-15pt}
	\label{fig:teaser}
\end{figure}

Part of the reason for this lack of success is probably that current
datasets used for training 
are either too small or too noisy 
(we return to this point below in related work).
Compared to image classification, action classification in video
has the additional challenge of variations in motion and viewpoint, and
so might be expected to require {\em more} training examples than that
of ImageNet (1000 per class) -- yet UCF-101 has only 100 examples per
class.
Another important reason is that current ConvNet
architectures are not able to take full advantage of temporal
information and their performance is consequently 
often dominated by spatial (appearance) recognition.

As can be seen from Fig.~\ref{fig:teaser}, some actions can be
identified from a still image from their appearance alone (archery in
this case).  For others, though, individual frames can be ambiguous,
and motion cues are necessary. Consider, for example, discriminating
walking from running, yawning from laughing, 
or in swimming, crawl from breast-stroke. 
The two-stream architecture \cite{Simonyan14b}
incorporates motion information by training separate ConvNets for 
both appearance
in still
images and stacks of optical flow. Indeed, this work showed
that optical flow information alone was sufficient to discriminate
most of the actions in UCF101.

Nevertheless, the
two-stream architecture (or any previous method) 
is not able to exploit two very important
cues for action recognition in video: 
(i) recognizing what is moving where, i.e.\ registering appearance
recognition (spatial cue) with optical flow recognition (temporal cue);
and (ii) how these cues evolve over time. 

Our objective in this paper is to rectify this by developing an
architecture that is able to fuse spatial and temporal cues at several
levels of granularity in feature abstraction, and with spatial as well
as temporal integration. In particular, Sec.~\ref{sec:approach} investigates three aspects of fusion: (i) in Sec.~\ref{sec:spatialFusion} {\em how} to fuse the two networks (spatial and temporal)
taking account of {\em spatial} registration?  (ii) in Sec.~\ref{sec:whereToFuse}  {\em where} to fuse
the two networks?  And, finally in Sec.~\ref{sec:temporalFusion} (iii) how to fuse the networks {\em temporally}?  In each of these investigations we select the optimum outcome (Sec.~\ref{sec:evaluation}) and then, putting this together, propose a novel architecture (Sec.~\ref{sec:architecture}) for spatiotemporal fusion of two stream networks that achieves  state of the art performance in Sec.~\ref{sec:sotaComparison}.

We implemented our approach using the MatConvNet toolbox \cite{vedaldi15matconvnet} and made our code publicly available at \href{https://github.com/feichtenhofer/twostreamfusion}{https://github.com/feichtenhofer/twostreamfusion}


\section{Related work}
\label{sec:related_work}
Several recent work on using ConvNets for action recognition in temporal
sequences have investigated the question of how to go beyond simply
using the framewise appearance information, and exploit the temporal
information.
A natural extension is to stack consecutive video frames and 
extend 2D ConvNets into time~\cite{JiPAMI13} so that the first layer learns
spatiotemporal features. \cite{Karpathy14} study 
several approaches for temporal sampling, including
early fusion (letting the first layer filters operate over frames as in~\cite{JiPAMI13}), slow fusion (consecutively increasing the temporal receptive field as the layers increase) and late fusion (merging fully connected layers of two separate networks that operate on temporally distant frames).
Their architecture is
not particularly sensitive to the temporal modelling, and 
they achieve similar levels of performance by
a purely spatial network, indicating that their model is not gaining much
from the temporal information.

The recently proposed C3D method \cite{C3DICCV2015} learns 3D ConvNets
on a limited temporal support of 16 consecutive frames with all filter kernels
of size 3$\times$3$\times$3.   They report better performance than \cite{Karpathy14} by letting all filters operate over space and time. However, their network is considerably deeper than \cite{JiPAMI13,Karpathy14} with a structure similar to the very deep networks in \cite{simonyan2014very}. Another way of learning spatiotemporal  relationships is proposed in \cite{Sun15}, where the authors factorize 3D convolution into a 2D spatial and a 1D temporal convolution. Specifically, their temporal convolution is a 2D convolution over time as well as the feature channels and is only performed at higher layers of the network.

\cite{ng2015beyond} compares 
several temporal feature pooling architectures to combine information
across longer time periods. They conclude that temporal pooling of
convolutional layers performs better than slow, local, or late
pooling, as well as temporal convolution. They also investigate
ordered sequence modelling by feeding the ConvNet features
into a recurrent network with Long Short-Term Memory (LSTM)
cells. Using LSTMs, however did not give an improvement over temporal
pooling of convolutional features.

The most closely related work to ours, and the one we extend here, 
is the two-stream ConvNet
architecture proposed in \cite{Simonyan14b}. The method first
decomposes video into spatial and temporal components by using RGB and
optical flow frames. These components are fed into separate deep
ConvNet architectures, to learn spatial as well as temporal
information about the appearance and movement of the objects in a
scene.  Each stream is performing video recognition on its own and for
final classification, softmax scores are combined by late fusion. The
authors compared several techniques to align the optical flow frames
and concluded that simple stacking of $L=10$ horizontal and vertical
flow fields performs best. They also employed multitask learning on
UCF101 and HMDB51 to increase the amount of training data and improve
the performance on both. To date, this method is the most effective
approach of applying deep learning to action recognition, especially
with limited training data. The two-stream approach has recently been employed into several action recognition methods~\cite{Cheron15, %
donahue2015long,gkioxari2014finding,ng2015beyond,srivastava2015unsupervised,Venugopalan15,Weinzaepfel15}.

Also related to our work is the bilinear method \cite{lin2015bilinear}
which correlates the output of two ConvNet layers by performing an
outer product at each location of the image. The resulting bilinear
feature is pooled across all locations into an orderless
descriptor. Note that this is closely related to second-order pooling
\cite{Carreira12} of hand-crafted SIFT features.

In terms of datasets, \cite{Karpathy14} introduced
the Sports-1M dataset which has a large number of
videos ($\approx$1M) and classes (487). However, the videos
are gathered automatically and therefore are not free of label noise. Another
large scale dataset is the THUMOS dataset
\cite{gorbanTHUMOS15} that has over 45M frames. Though, only a small
fraction of these actually contain the labelled action and thus are
useful for supervised feature learning. Due to the label noise,
learning spatiotemporal ConvNets still largely relies on smaller,
but temporally consistent datasets such as UCF101~\cite{UCF101} or
HMDB51~\cite{kuehne2011hmdb} which contain short videos of
actions. This facilitates learning, but comes with the risk of
severe overfitting to the training data.

\section{Approach} \label{sec:approach}
We build upon the the two-stream architecture in
\cite{Simonyan14b}. 
This architecture has two main drawbacks: (i) it is not able
to learn the pixel-wise
correspondences between spatial and temporal features (since
fusion is only on the classification scores), and (ii) it is limited in
temporal scale as the spatial ConvNet operates only on single frames
and the temporal ConvNet only on a stack of $L$ temporally adjacent
optical flow frames (\eg $L=10$). The implementation of~\cite{Simonyan14b} addressed the
latter problem to an extent by temporal pooling across regularly
spaced samples in the video, but this does not allow the modelling of
temporal evolution of actions.

\subsection{Spatial fusion} \label{sec:spatialFusion}

In this section we consider different architectures for fusing the two
stream networks. However, the same issues arise when spatially fusing any two
networks so are not tied to this particular application.

To be clear, our intention here is to fuse the two networks (at a
particular convolutional layer) such that channel responses at the
same pixel position are put in correspondence. To motivate this, consider
for example 
discriminating between the actions of brushing teeth and brushing hair.
If a hand
moves periodically 
at some spatial location then the temporal network can recognize
that motion, and the spatial network can recognize the location (teeth or hair)
and their combination then discriminates the action.

This spatial correspondence is easily achieved when the two networks
have the same spatial resolution at the layers to be fused, simply by
overlaying (stacking) layers from one network on the other (we make
this precise below).  However, there is also the issue of which {\em
	channel} (or channels) in one network {\em corresponds} to the {\em
	channel} (or channels) of the other network.

Suppose for the moment that different channels in the spatial network
are responsible for different facial areas (mouth, hair, etc), and one
channel in the temporal network is responsible for periodic motion
fields of this type. Then, after the channels are stacked, the filters
in the subsequent layers must learn the correspondence between these
appropriate channels (\eg as weights in a convolution filter) in order
to best discriminate between these actions.

To make this more concrete, we now discuss a number of ways of fusing layers
between two networks, and for each describe the consequences in terms of
correspondence.

A fusion function $f:\bx^a_t, \bx^b_t, \rightarrow \by_t$ fuses two
feature maps $\bx^a_t \in \mathbb{R}^{H\times W\times D}$ and $\bx^b_t
\in \mathbb{R}^{H'\times W'\times D'}$, 
at time $t$, to produce an
output map $\by_t \in \mathbb{R}^{H''\times W''\times D''}$, where
$W,H$ and $D$ are the width, height and number of channels of the
respective feature maps. When applied to feedforward ConvNet
architectures, consisting of convolutional, fully-connected, pooling
and nonlinearity layers, $f$ can be applied at different points in the
network to implement \eg early-fusion, late-fusion or multiple
layer fusion. Various fusion functions $f$ can be used. We investigate the
following ones in this paper, and,
for simplicity, assume that $H = H' = H''$, $W = W'=W''$,
$D = D'$, and also drop the $t$ subscript.

\textbf{Sum fusion.} $\by^\text{sum} = f^\text{sum}(\bx^{a},\bx^{b})$ computes the sum of the two feature maps at the same spatial locations $i,j$ and feature channels $d$: 
\begin{align}
y^\text{sum}_{i,j,d}  = x^a_{i,j,d} + x^b_{i,j,d},
\label{eq:sum_fusion}
\end{align}
where $1\leq i \leq H, 1 \leq j \leq W, 1 \leq d \leq D$ and $\bx^a, \bx^b, \by \in \mathbb{R}^{H\times W\times D}$

Since the channel numbering is arbitrary, sum fusion simply defines an
arbitrary correspondence between the networks. Of course, subsequent learning
can employ this arbitrary correspondence to its best effect,
optimizing over the filters of each network to make this correspondence
useful.

\textbf{Max fusion.} $\by^{\max} = f^{\max}(\bx^{a},\bx^{b})$ similarly takes the maximum of the two feature map: 
\begin{align}
y^{\max}_{i,j,d}  = \max\{x^a_{i,j,d}, x^b_{i,j,d}\},
\end{align}
where all other variables are defined as above \eqref{eq:sum_fusion}.

Similarly to sum fusion, the correspondence between network channels is again
arbitrary.

\textbf{Concatenation fusion.} $\by^\text{cat} = f^\text{cat}(\bx^{a},\bx^{b})$ stacks the two feature maps at the same spatial locations $i,j$ across the feature channels $d$: 
\begin{align}
y^\text{cat}_{i,j,2d}  = x^a_{i,j,d} \qquad 
y^\text{cat}_{i,j,2d-1}  = x^b_{i,j,d},
\label{eq:cat_fusion}
\end{align}
where $\by \in \mathbb{R}^{H \times W \times 2D}$. 

Concatenation does not define a correspondence, but leaves this to subsequent 
layers to define (by learning suitable filters that weight the layers), as we
illustrate next.

\textbf{Conv fusion.} $\by^\text{conv} = f^\text{conv}(\bx^{a},\bx^{b})$ first stacks the two feature maps at the same spatial locations $i,j$ across the feature channels $d$ as above \eqref{eq:cat_fusion} and subsequently convolves the stacked data with a bank of filters $\bff\in\real^{1\times 1\times 2D\times D}$ and biases $b\in\real^{D}$
\begin{align}
\by^\text{conv}  = \by^\text{cat} *\bff + b,
\label{eq:conv_fusion}
\end{align}
where the number of output channels is $D$, and the filter has dimensions
$1 \times 1 \times 2D$.
Here, the filter $\bff$ is used to reduce the dimensionality by a
factor of two and is able to model weighted combinations of the two
feature maps $\bx^{a},\bx^{b}$ at the same spatial (pixel) location. When used
as a trainable filter kernel in the network, $\bff$ is able to {\em
	learn} correspondences of the two feature maps that minimize a joint
loss function. For example, if $\bff$ is learnt to be the
concatenation of two permuted identity matrices
$\bone'\in\real^{1\times 1\times D\times D}$, then the $i$th channel
of the one network is only combined with the $i$th channel of the other (via summation).  

Note that if there is no dimensionality reducing conv-layer injected
after concatenation, the number of input channels of the upcoming
layer is $2 D$.

\textbf{Bilinear fusion.} $\by^\text{bil} = f^\text{bil}(\bx^{a},\bx^{b})$ computes a matrix outer product of the two features at each pixel location, followed by a summation over the locations:
\begin{align}
\by^\text{bil} = \sum_{i=1}^H \sum_{j=1}^W \bx^{a\top}_{i,j}  \bx^{b}_{i,j}.
\end{align}
The resulting feature $\by^\text{bil} \in \mathbb{R}^{D^{2}}$ captures
multiplicative interactions at corresponding spatial locations. The
main drawback of this feature is its high dimensionality. To make
bilinear features usable in practice, it is usually applied at ReLU5, the fully-connected layers
are removed \cite{lin2015bilinear} and power- and $L2$-normalisation is
applied for effective classification with linear SVMs.

The advantage of bilinear fusion is that every channel of one network
is combined (as a product) with every channel of the other
network. However, the disadvantage is that all spatial information is
marginalized out at this point.

\textbf{Discussion:} These operations illustrate a range of possible fusion methods. Others could be considered, for example: taking the pixel wise
product of channels (instead of their sum or max), or the (factorized) outer product without sum pooling across locations \cite{Oh15}. 

Injecting fusion layers can have significant impact on the number of parameters
and layers in a two-stream network, especially if only the network
which is fused into is kept and the other network tower is truncated, 
as illustrated in Fig.~\ref{fig:where_to_fuse} (left).
Table \ref{tab:fusion_comparisons} shows how the number of layers and
parameters are affected by different fusion methods for the case of
two VGG-M-2048 models (used in \cite{Simonyan14b}) containing five
convolution layers followed by three fully-connected layers each.
Max-, Sum and Conv-fusion at ReLU5 (after the last convolutional layer)
removes nearly {\em half} of the parameters in the
architecture as only one tower of fully-connected layers is used after
fusion. Conv fusion has slightly more parameters (97.58M) compared to
sum and max fusion (97.31M) due to the additional filter that is used
for channel-wise fusion and dimensionality reduction. Many more
parameters are involved in concatenation fusion, which does not
involve dimensionality reduction after fusion and therefore doubles
the number of parameters in the first fully connected layer. In
comparison, sum-fusion at the softmax layer requires all layers (16)
and parameters (181.4M) of the two towers.

In the experimental section (Sec.~\ref{sec:eval_ts_fusion}) we
evaluate and compare the performance of each of these possible fusion
methods in terms of their classification accuracy.

\subsection{Where to fuse the networks} \label{sec:whereToFuse}
\begin{figure}[!h]
	\centering
	\resizebox {0.45\textwidth }{!}{ 
		\includegraphics[width=1\textwidth]	{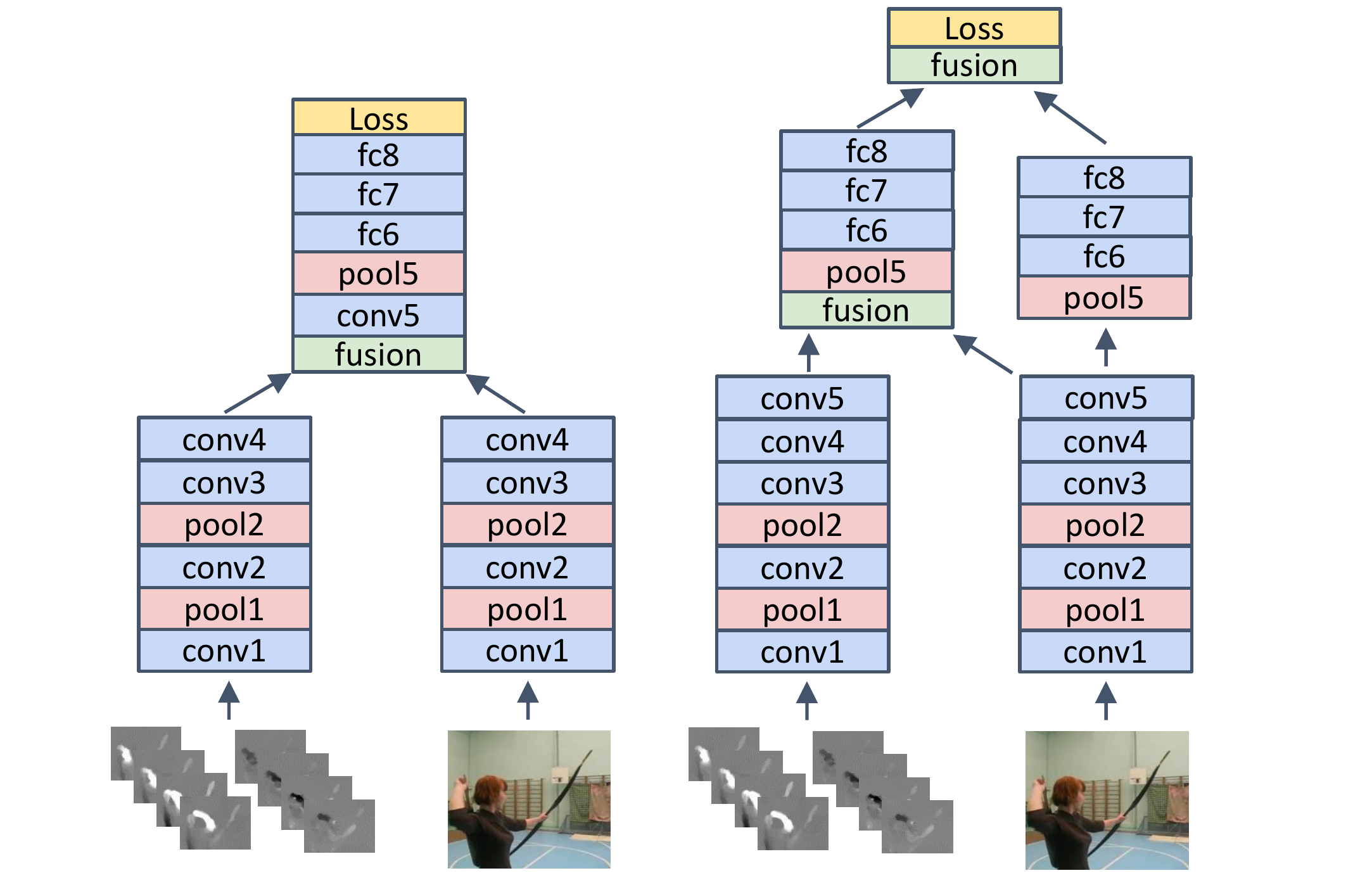}
	}
	\vspace{-5pt}
	\caption{Two examples of where a fusion layer can be placed. The left example
		shows fusion after the fourth conv-layer. Only a single network tower
		is used from the point of fusion.
		The right figure shows fusion at two layers (after conv5
		and after fc8) where both network towers are kept, one as a hybrid
		spatiotemporal net and one as a purely spatial network.  }
	\label{fig:where_to_fuse}
	\vspace{-15pt}
\end{figure}

As noted above, fusion can be applied at any point in the two
networks, with the only constraint that the two input maps $\bx^a_t
\in \mathbb{R}^{H\times W\times D}$ and $\bx^b_t
\in \mathbb{R}^{H'\times W'\times D}$, 
at time $t$, have the same spatial dimensions; \ie $H = H'$, $W = W'$.
This can be achieved by using an ``upconvolutional'' layer \cite{Zeiler13}, or if the dimensions are similar, upsampling can be achieved by padding the smaller map with zeros. 

Table \ref{tab:fusion_layer_comparisons} compares the number of
parameters for fusion at different layers in the two networks for the
case of a VGG-M model. Fusing after different conv-layers has
roughly the same impact on the number of parameters, as most of these
are stored in the fully-connected layers. 
Two networks can also be fused at two layers, as illustrated 
in Fig.~\ref{fig:where_to_fuse} (right). This achieves the original 
objective of pixel-wise registration of the channels from each network
(at conv5) but does not lead to a reduction in the number of parameters
(by half if fused only at conv5, for example).
In the experimental section (Sec.~\ref{sec:eval_ts_fusion_layer}) we evaluate and compare
both the performance of fusing at different levels, and fusing at
multiple layers simultaneously. 

\begin{figure}[!h]
	\centering
	\resizebox {0.5\textwidth }{!}{ 
		\includegraphics[width=1\textwidth]	{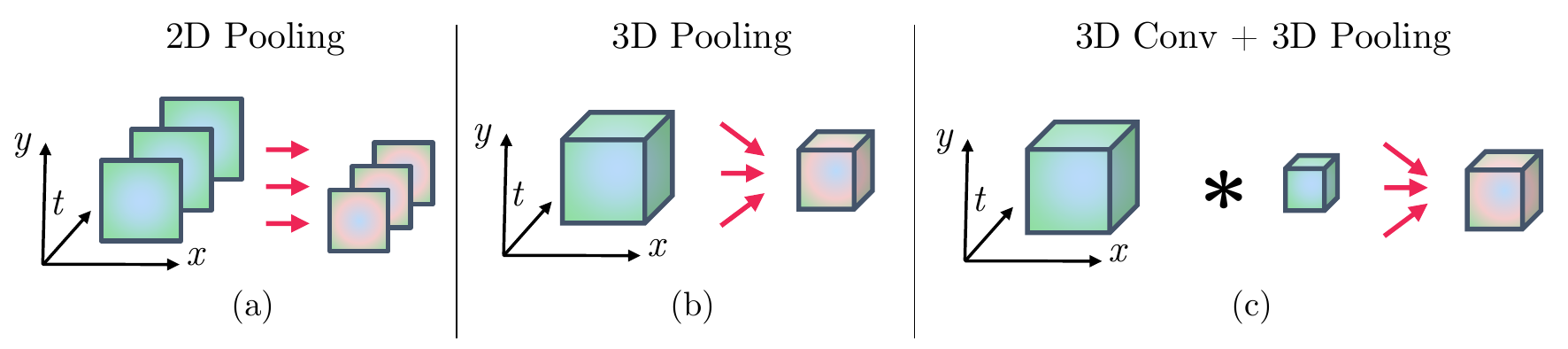}
	}
	\caption{ Different ways of fusing temporal information. (a) 2D pooling ignores time and simply pools over spatial neighbourhoods to individually shrink the size of the feature maps  for each temporal sample. (b) 3D pooling pools from local spatiotemporal neighbourhoods by first stacking the feature maps across time and then shrinking this spatiotemporal cube. (c) 3D conv + 3D pooling additionally performs a convolution with a fusion kernel that spans the feature channels, space and time before 3D pooling.
	}
	\label{fig:temporal_fusion}
	\vspace{-15pt}
\end{figure}
\subsection{Temporal fusion} \label{sec:temporalFusion}

\begin{figure*}[!t]
	\centering
	\resizebox {0.89\textwidth }{!}{ 
		\includegraphics[width=1\textwidth]	{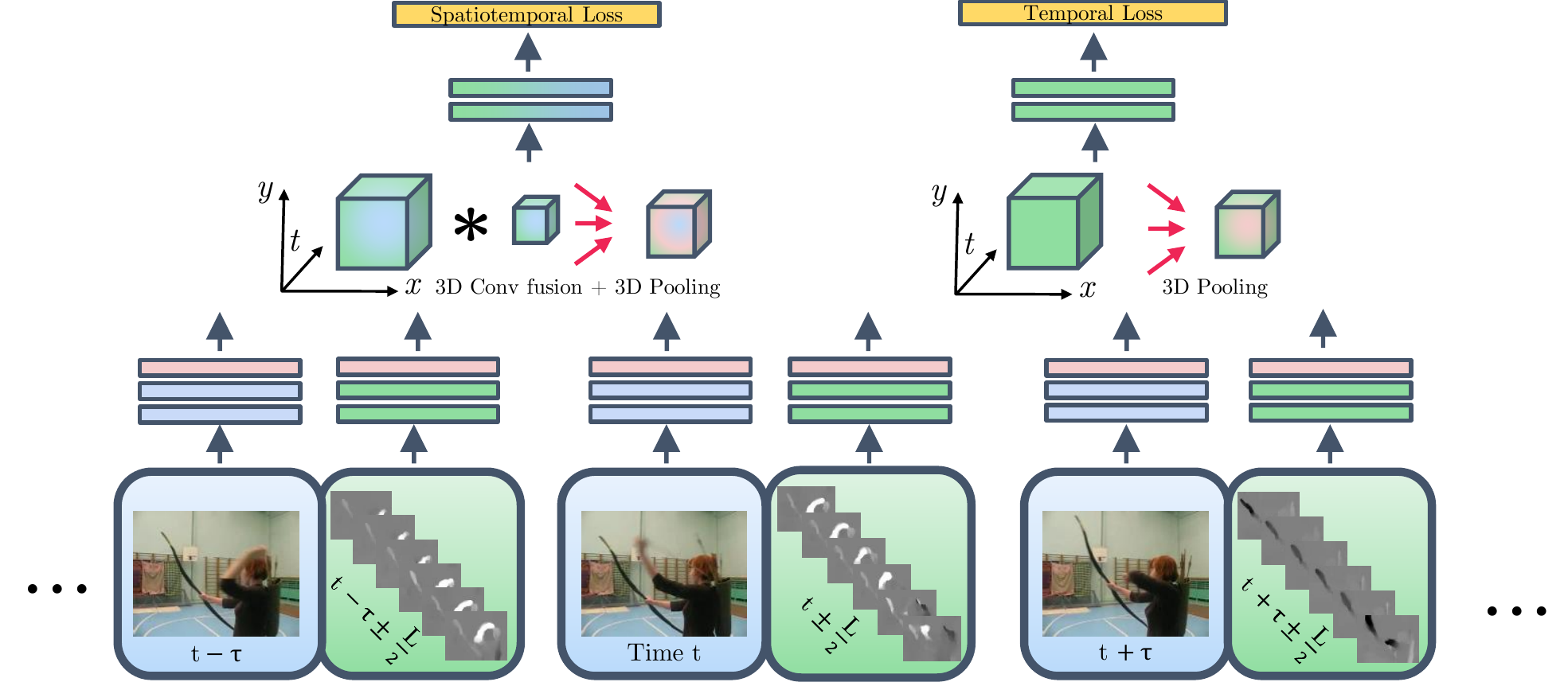}
	}
	\caption{Our spatiotemporal fusion ConvNet applies two-stream ConvNets, that capture short-term information at a fine temporal scale ($t \pm \frac{L}{2}$), to temporally adjacent inputs at a coarse temporal scale ($t + T \tau$). The two streams are fused by a 3D filter that is able to learn correspondences between highly abstract features of the spatial stream (blue) and temporal stream (green), as well as local weighted combinations in $x,y,t$. The resulting features from the fusion stream and the temporal stream are 3D-pooled in space and time to learn spatiotemporal (top left) and purely temporal (top right) features for recognising the input video.
	}
	\label{fig:architecture}
	\vspace{-10pt}
\end{figure*}

We now consider techniques to combine feature maps $\bx_t$ over time
$t$, to produce an output map $\by_t$. One way of processing temporal inputs is by averaging the network predictions over time (as used in~\cite{Simonyan14b}). In that case the architecture only pools in 2D ($xy$); see Fig.~\ref{fig:temporal_fusion}(a).

Now consider the input of a temporal pooling layer as feature maps  $\bx \in \mathbb{R}^{H\times W \times T \times D}$ which are generated by stacking spatial maps across time $t=1\ldots T$.

\textbf{3D Pooling:} 
applies max-pooling to the stacked data within a 3D pooling cube of
size $W' \times H' \times T'$. This is a straightforward extension of
2D pooling to the temporal domain, as illustrated in
Fig.~\ref{fig:temporal_fusion}(b). For example, if three temporal samples
are pooled, then a $3 \times 3 \times 3$ max pooling could be used
across the three stacked corresponding channels. Note, there is no
pooling {\em across} different channels.

\textbf{3D Conv + Pooling:} first convolves the four dimensional input  $\bx$ with a bank of $D'$ filters $\bff\in\real^{W'' \times H'' \times T'' \times D\times D'}$ and biases $b\in\real^{D}$ 
\begin{align}
\by  = \bx_t *\bff + b,
\label{eq:3Dconv_fusion}
\end{align}
as \eg in~\cite{C3DICCV2015}, followed by 3D
pooling as described above. This
method is illustrated in Fig.~\ref{fig:temporal_fusion}(c).
The filters
$\bff$ are able to model weighted combinations of the features in
a local spatio-temporal neighborhood using kernels of size
$W'' \times H'' \times T'' \times D$. Typically the neighborhood
is $3 \times 3 \times 3$ (spatial $\times$ temporal).

\textbf{Discussion.} The authors of \cite{ng2015beyond}
evaluate several additional methods to combine two-stream ConvNets
over time. They find temporal max-pooling of convolutional layers
among the top performers. We generalize max-pooling here to 3D pooling that provides invariance to small changes of the features' position over time.
Further, 3D conv allows spatio-temporal filters to be learnt~\cite{C3DICCV2015,Taylor10}.
For example, the filter could learn to center weight the central temporal
sample, or to differentiate in time or space.

\subsection{Proposed architecture}  \label{sec:architecture}

We now bring together the ideas from the previous sections to propose
a new spatio-temporal fusion architecture and motivate our choices
based on our empirical evaluation in Sec.~\ref{sec:evaluation}. The
choice of the spatial fusion method, layer and temporal fusion is based
on the experiments in sections~\ref{sec:eval_ts_fusion},
~\ref{sec:eval_ts_fusion_layer} and~\ref{sec:fusing_over_space_time},
respectively.

Our proposed architecture (shown in Fig.~\ref{fig:architecture}) can
be viewed as an extension of the architecture in
Fig.~\ref{fig:where_to_fuse} (left) over time. We fuse the two
networks, at the last convolutional layer (after ReLU) {\em into} the
spatial stream to convert it into a spatiotemporal stream by using 3D
Conv fusion followed by 3D pooling (see Fig.~\ref{fig:architecture},
left).  Moreover, we {\em do not truncate} the temporal stream and
also perform 3D Pooling in the temporal network (see
Fig.~\ref{fig:architecture}, right). The losses of both streams are
used for training and during testing we average the predictions of the
two streams. In our empirical evaluation
(Sec.~\ref{sec:sotaComparison}) we show that keeping both streams
performs slightly better than truncating the temporal stream after
fusion.

Having discussed how to fuse networks over time, we discuss
here the issue of how often to sample the temporal sequence.
The temporal fusion layer receives $T$ temporal chunks that are $\tau$ frames apart; \ie the two stream towers are applied to the input video at time $t, t+\tau,
\ldots t+T\tau$.
As shown in Fig.~\ref{fig:architecture} this enables us to capture short scale ($t\pm \frac{L}{2}$) temporal features at the input of the temporal network (\eg the drawing of an arrow) and put
them into context over a longer temporal scale ($t+T\tau$) at a higher
layer of the network (\eg drawing an arrow, bending a bow, and
shooting an arrow).  

Since the optical flow stream has a temporal receptive field of $L=10$ frames, the architecture operates on a total temporal receptive field of $T\times L$. 
Note that $\tau < L$ results in overlapping inputs for the temporal stream, whereas $\tau \geq L$ produces temporally non-overlapping features.

After fusion, we let the 3D pooling operate on $T$ spatial feature maps
that are $\tau$ frames apart. As features may change their spatial
position over time, combining spatial and temporal pooling to 3D
pooling makes sense. For example, the output of a VGG-M network at conv5 has
an input stride of $16$ pixels and captures high level features from a
receptive field of $139\times 139$ pixels. Spatiotemporal pooling
of conv5 maps that are $\tau$ frames distant in time can therefore capture
features of the same object, even if they slightly move.

\subsection{Implementation details} \label{sec:implementation details}

\textbf{Two-Stream architecture.} We employ two pre-trained ImageNet models. First, for sake of comparison to the original two-stream approach \cite{Simonyan14b}, the VGG-M-2048 model \cite{chatfield2014devil} with 5 convolutional and 3 fully-connected layers. Second, the very deep VGG-16 model \cite{simonyan2014very} that has 13 convolutional and 3 fully-connected layers. 
We first separately train the two streams as described in \cite{Simonyan14b}, but with some subtle differences: We do not use RGB colour jittering; Instead of decreasing the learning rate according to a fixed schedule, we lower it after the validation error saturates; For training the spatial network we use lower dropout ratios of 0.85 for the first two fully-connected layers. Even lower dropout ratios (up to 0.5) did not decrease performance significantly. 

For the temporal net, we use optical flow \cite{BroxECCV2004,Zach2007_OpticalFlow} stacking with $L=10$ frames \cite{Simonyan14b}. We also initialised the temporal net with a model pre-trained on ImageNet, since this generally facilitates training speed without a decrease in performance compared to our model trained from scratch. The network input is rescaled beforehand, so that the smallest side of the frame equals 256. We also pre-compute the optical flow before training and store the flow fields as JPEG images (with clipping of displacement vectors larger than 20 pixels).  We do not use batch normalization \cite{ioffe2015batch}.

\textbf{Two-Stream ConvNet fusion.} For fusion, these networks are finetuned with a batch size of $96$ and a learning rate starting from $10^{-3}$ which is reduced by a factor of $10$ as soon as the validation accuracy saturates. We only propagate back to the injected fusion layer, since full backpropagation did not result in an improvement.
In our experiments we only fuse between layers with the same output resolution; except for fusing a VGG-16 model at ReLU5\_3 with a VGG-M model at ReLU5, where we pad the slightly smaller output of VGG-M ($13\times 13 $, compared to $ 14\times 14 $) with a row and a column of zeros.
For Conv fusion, we found that careful initialisation of the injected fusion layer (as in (\ref{eq:conv_fusion})) is very important. We compared several methods and found that initialisation
by identity matrices (to sum the two networks) performs as well as random initialisation. 

\textbf{Spatiotemporal architecture.}
For our final architecture described in Sec.~\ref{sec:architecture},
the 3D Conv fusion kernel $\bff$ has dimension $3 \times 3 \times 3 \times
1024 \times 512$ and $T=5$, i.e.\ the spatio-temporal filter has dimension
$H''\times W''\times T'' = 3\times 3\times 3$, the $D= 1024$ results
from concatenating the ReLU5 from the spatial and temporal streams,
and the $D' = 512$ matches the number of input channels of the following FC6 layer. 

The 3D Conv filters are also initialised by stacking two identity
matrices for mapping the $1024$ feature channels to $512$.  Since the
activations of the temporal ConvNet at the last convolutional layer are
roughly 3 times lower than its appearance counterpart, we initialise
the temporal identity matrix of $\bff$ by a factor of 3 higher. 
The spatiotemporal part of $\bff$
is initialised using a Gaussian of
size $3\times 3\times 3$ and $\sigma=\bone$.
Further, we do not fuse at the prediction layer during training, as
this would bias the loss towards the temporal architecture, because
the spatiotemporal architecture requires longer to adapt to the fused
features. 

Training 3D ConvNets is
even more prone to overfitting than the two-stream ConvNet fusion, 
and requires additional augmentation as follows.
During finetuning, at each 
training iteration we sample the $T= 5$ frames from each of
the 96 videos in a batch by randomly sampling 
the starting frame, and then randomly sampling the temporal stride ($\tau$) 
$ \in \left [ 1,10 \right ] $ (so 
operating over a total of between  $15$ and $50$ frames). 
Instead of cropping a fixed sized $224 \times 224$ input patch, we randomly jitter its width and height by $\pm 25\%$ and
rescale it to $224 \times 224$. The rescaling is chosen randomly and
may change the aspect-ratio. 
Patches are only cropped at a maximum of $25\%$
distance from the image borders (relative to the width and height).
Note, the position (and size, scale, horizontal flipping) of the crop is 
randomly selected in the first frame (of a multiple-frame-stack) and then
the same spatial crop is applied to all frames in the stack.

\textbf{Testing.} Unless otherwise specified, only the $T=5$ frames (and their
horizontal flips) are sampled, compared to the 25 frames in \cite{Simonyan14b}, to
foster fast empirical evaluation. In addition we employ fully convolutional testing where the entire frame is used (rather than
spatial crops).

	\section{Evaluation} \label{sec:evaluation}
\subsection{Datasets and experimental protocols}
We evaluate our approach on two popular action recognition datasets. First, UCF101 \cite{UCF101}, which consists of 13320 action videos in 101 categories. 
The second dataset is HMDB51 \cite{kuehne2011hmdb}, which contains 6766 videos that have been annotated for 51 actions. 
For both datasets, we use the provided evaluation protocol and report the mean average accuracy over the three splits into training and test data. 

\subsection{How to fuse the two streams spatially?}

\label{sec:eval_ts_fusion}
For these experiments we use the same network architecture as in~\cite{Simonyan14b}; i.e. two VGG-M-2048 nets \cite{chatfield2014devil}. The fusion layer is injected at the last convolutional layer, after rectification, \ie its input is the output of ReLU5 from the two streams. This is chosen because, in preliminary experiments, it provided better results than alternatives such as the non-rectified output of conv5. At that point the features are already highly informative while still providing coarse location information. 
After the fusion layer a single processing stream is used.

We compare different fusion strategies in Table \ref{tab:fusion_comparisons} where we report the average accuracy on the first split of UCF101. We first observe that our performance for softmax averaging (85.94\%) compares favourably to the one reported in \cite{Simonyan14b}. Second we see that Max and Concatenation perform considerably lower than Sum and Conv fusion. Conv fusion performs best and is slightly better than Bilinear fusion and simple fusion via summation. For the reported Conv-fusion result, the convolution kernel $\bff$ is initialised by identity matrices that perform summation of the two feature maps. Initialisation via random Gaussian noise ends up at a similar performance 85.59\% compared to identity matrices (85.96\%), however, at a much longer training time. This is interesting, since this, as well as the high result of Sum-fusion, suggest that simply summing the feature maps is already a good fusion technique and learning a randomly initialised combination does not lead to significantly different/better results.

For all the fusion methods shown in Table \ref{tab:fusion_comparisons}, fusion  at FC layers results in lower performance compared to ReLU5, with the ordering of the methods being the same as in Table \ref{tab:fusion_comparisons}, except for bilinear fusion which is not possible at FC layers. Among all FC layers, FC8 performs better than FC7 and FC6, with Conv fusion at 85.9\%, followed by Sum fusion at 85.1\%.  We think the reason for ReLU5 performing slightly better is that at this layer spatial correspondences between appearance and motion are fused, which would have already been collapsed at the FC 
layers~\cite{Mahendran15}.

\begin{table}[!t]
	\begin{center}
		\resizebox{0.5\textwidth}{!}{
			\begin{tabular}{|l|c|c|c|c|}
				\hline
				Fusion Method & Fusion Layer & Acc. & \#layers & \#parameters \\
				\hline
				Sum \cite{Simonyan14b} & Softmax  &  85.6\% & 16 & 181.42M \\
				Sum (ours) & Softmax & 85.94\% & 16 & 181.42M \\
				\hline
				Max & ReLU5 & 82.70\% & 13 & 97.31M\\	
				Concatenation & ReLU5 & 83.53\% & 13 &  172.81M \\	
				Bilinear \cite{lin2015bilinear} & ReLU5 & 85.05\% & 10 & 6.61M+SVM \\								
				Sum & ReLU5 & 85.20\% & 13 & 97.31M\\ 
				Conv   & ReLU5 & 85.96\% & 14 & 97.58M \\
				\hline
			\end{tabular}
		}
		
		\caption{ Performance comparison of different spatial fusion strategies (Sec.~\ref{sec:spatialFusion}) on UCF101 (split 1). Sum fusion at the softmax layer corresponds to averaging the two networks predictions and therefore includes the parameters of both 8-layer VGG-M models. Performing fusion at ReLU5 using Conv or Sum fusion does not significantly lower classification accuracy. Moreover, this requires
			only half of the parameters in the softmax fusion network. Concatenation has lower performance and requires twice as many parameters in the FC6 layer (as Conv or Sum fusion).
			Only the bilinear combination enjoys much fewer parameters as there are no FC layers involved; however, it has to employ an SVM to perform comparably.  }
		\label{tab:fusion_comparisons}
	\end{center}
	\vspace{-20pt}
\end{table}

\subsection{Where to fuse the two streams spatially?}
\label{sec:eval_ts_fusion_layer}

Fusion from different layers is compared in Table \ref{tab:fusion_layer_comparisons}.  Conv fusion is used and the  fusion layers are initialised by an identity matrix that sums the
activations from previous layers. 
Interestingly, fusing and 
truncating one net at ReLU5 achieves around the same classification accuracy on
the first split of UCF101 (85.96\% vs 86.04\%) as an additional fusion
at the prediction layer (FC8), but at a much lower number of total
parameters (97.57M vs 181.68M). Fig.~\ref{fig:where_to_fuse}
shows how these two examples are implemented.

\begin{table}
	\begin{center}
		\resizebox{0.5\textwidth}{!}{
			\begin{tabular}{|l|c|c|c|}
				\hline
				Fusion Layers & Accuracy & \#layers & \#parameters \\
				\hline
				ReLU2  & 82.25\%  & 11 & 91.90M\\
				ReLU3  & 83.43\%  & 12& 93.08M\\
				ReLU4  & 82.55\% & 13& 95.48M \\
				ReLU5  & 85.96\%  & 14 &	 97.57M\\
				ReLU5 + FC8  & 86.04\%  & 17 & 181,68M \\
				ReLU3 + ReLU5 + FC6   & 81.55\% & 17 &	 190,06M\\
				\hline
			\end{tabular}
		}
		\caption{Performance comparison for Conv fusion \eqref{eq:conv_fusion} at different fusion layers. An earlier fusion (than after conv5) results in weaker performance. Multiple fusions also lower performance if early layers are incorporated (last row). Best performance is achieved for fusing at ReLU5 or at ReLU5+FC8 (but with nearly double the parameters involved).  }
		\label{tab:fusion_layer_comparisons}
	\end{center}
	\vspace{-20pt}
\end{table}

\subsection{Going from deep to very deep models} \label{sec:deep_to_vd}
For computational complexity reasons, all previous experiments were performed with two VGG-M-2048 networks (as in \cite{Simonyan14b}). Using deeper models, such as the very deep networks in \cite{simonyan2014very} can, however, lead to even better performance in image recognition tasks \cite{cimpoi2015deep,lin2015bilinear,szegedy2014going}. Following that, we train a 16 layer network, VGG-16, \cite{simonyan2014very} on UCF101 and HMDB51. All models are pretrained on ImageNet and separately trained for the target dataset, except for the temporal HMDB51 networks which are initialised from the temporal UCF101 models. For VGG-16, we use TV-L1 optical flow \cite{Zach2007_OpticalFlow} and apply a similar augmentation technique as for 3D ConvNet training (described in Sec.~\ref{sec:implementation details}) that samples from the image corners and  its centre \cite{wang2015towards}. The learning rate is set to $50^{-4}$ and decreased by a factor of $10$ as soon as the validation objective saturates. 

\begin{table}
	\begin{center}
		\resizebox{0.5\textwidth}{!}{
			\begin{tabular}{|c|cc|cc|}
				\cline{2-5}
				\multicolumn{1}{c}{} &	\multicolumn{2}{|c|}{UCF101 (split 1)} & \multicolumn{2}{c|}{HMDB51 (split 1)}  \\ \hline
				Model &  VGG-M-2048 & VGG-16  & VGG-M-2048 & VGG-16   \\ \hline
				Spatial & 74.22\% & 82.61\%  & 36.77\% &  47.06\%  \\
				Temporal & 82.34\% & 86.25\% &  51.50\% & 55.23\% \\
				Late Fusion & 85.94\% & 90.62\% & 54.90\% & 58.17\% \\
				\hline
			\end{tabular}
		}
		\caption{Performance comparison of deep (VGG-M-2048) vs. very deep (VGG-16) Two-Stream ConvNets on the UCF101 (split1) and HMDB51 (split1). Late fusion is implemented by averaging the prediction layer outputs. Using deeper networks boosts performance at the cost of computation time. }
		\label{tab:two_stream_performance}
	\end{center}
	\vspace{-20pt}
\end{table}
The comparison between deep and very deep models is shown in Table \ref{tab:two_stream_performance}. On both datasets, one observes that going to a deeper spatial model boosts performance significantly (8.11\% and 10.29\%), whereas a deeper temporal network yields a lower accuracy gain (3.91\% and 3.73\%).

\subsection{How to fuse the two streams temporally?}

\label{sec:fusing_over_space_time}

\begin{table}
	\begin{center}
		\resizebox{0.45\textwidth}{!}{
			\begin{tabular}{|c|c|c|c|c|}
				\hline
				Fusion  & \multirow{ 2}{*}{Pooling}  & Fusion & \multirow{ 2}{*}{UCF101}  & \multirow{ 2}{*}{HMDB51}  \\
				Method &   &  Layers&  &  \\
				
				\hline
				2D Conv &  2D  & ReLU5 + &  89.35\% & 56.93\%  \\
				2D Conv &  3D  & ReLU5 + &  89.64\% & 57.58\% \\
				3D Conv &  3D  & ReLU5 + & 90.40\% & 58.63\%  \\	
				
				\hline
			\end{tabular}
		}
		\caption{Spatiotemporal two-stream fusion on UCF101 (split1) and HMDB51 (split1). The models used are VGG-16 (spatial net) and VGG-M (temporal net).  The ``+'' after a fusion layer indicates that both networks and their loss are kept after fusing, as this performs better than truncating one network. Specifically, at ReLU5 we fuse from the temporal net into the spatial network, then perform either 2D or 3D pooling at Pool5 and compute a loss for each tower. During testing, we average the FC8 predictions for both towers. }
		\label{tab:fusion_time}
		\vspace{-20pt}
	\end{center}
\end{table}

Different temporal fusion strategies are shown in Table~\ref{tab:fusion_time}.
In the first row of Table~\ref{tab:fusion_time} we observe that conv fusion performs better than averaging the softmax output (\textit{cf.~}Table \ref{tab:two_stream_performance}). Next, we find that applying 3D pooling instead of using 2D pooling after the fusion layer increases performance on both datasets, with larger gains on HMDB51. Finally, the last row of Table~\ref{tab:fusion_time} lists results for applying a 3D filter for fusion which further boosts recognition rates.

\subsection{Comparison with the state-of-the-art} \label{sec:sotaComparison}
Finally, we compare against the state-of-the-art over all three splits of UCF101 and HMDB51 in Table \ref{tab:SOTA}. We use the same method as shown above, \ie fusion by 3D Conv and 3D Pooling (illustrated in Fig.~\ref{fig:architecture}). 
For testing we average 20 temporal predictions from each network by densely sampling the input-frame-stacks and their horizontal flips. One interesting comparison is to the original two-stream approach \cite{Simonyan14b}, we improve by 3\% on UCF101 and HMDB51 by using a VGG-16 spatial (S) network and a VGG-M temporal (T) model, as well as by 4.5\% (UCF) and 6\% (HMDB) when using VGG-16 for both streams. 
Another interesting comparison is against the two-stream network in \cite{ng2015beyond}, which employs temporal conv-pooling after the last dimensionality reduction layer of a GoogLeNet \cite{szegedy2014going} architecture. They report 88.2\% on UCF101 when pooling over 120 frames and 88.6\% when using an LSTM for pooling. Here, our result of
92.5\% clearly underlines the importance of our proposed approach. 
Note also that using a single stream after 
temporal fusion achieves 91.8\%, compared to maintaining two streams and achieving 
92.5\%, but with far fewer parameters and a simpler architecture.

As a final experiment, we explore what benefit results from a late fusion of hand-crafted IDT features \cite{wangICCV13} with our representation. We simply average the SVM scores of the FV-encoded IDT descriptors (\ie HOG, HOF, MBH) with the predictions (taken before softmax) of our ConvNet representations. The resulting performance is shown in Table~\ref{tab:SOTA_IDT}. We achieve 93.5\% on UCF101 and 69.2\% HMDB51. This state-of-the-art result illustrates that there is still a degree of complementary between hand-crafted representations and our end-to-end learned ConvNet approach. 

\begin{table}
	\begin{center}
		\resizebox {0.45\textwidth }{!}{ 
			\begin{tabular}{|l|c|c|}
				\hline
				Method  &  UCF101 & HMDB51    \\
				\hline
				Spatiotemporal ConvNet \cite{Karpathy14} &  65.4\%  & -  \\       
				LRCN \cite{donahue2015long} & 82.9\% & - \\
				Composite LSTM Model \cite{srivastava2015unsupervised}  & 84.3\% & 44.0\\             
				C3D \cite{C3DICCV2015} & 85.2\% & - \\ 
				Two-Stream ConvNet (VGG-M) \cite{Simonyan14b} &	88.0\%  &	59.4\%  \\
				Factorized ConvNet \cite{Sun15} &	88.1\%    &	59.1\%  \\
				Two-Stream Conv Pooling \cite{ng2015beyond} &	88.2\%  &	-  \\
				Two-Stream ConvNet (VGG-16, \cite{wang2015towards}) &	91.4\%  &	58.5\%  \\
				Two-Stream ConvNet (VGG-16, ours) &	91.7\%  &	58.7\%  \\
				\hline
				Ours (S:VGG-16, T:VGG-M) & 90.8\% & 62.1\% \\
				Ours (S:VGG-16, T:VGG-16, & \multirow{ 2}{*}{91.8\%} & \multirow{ 2}{*}{64.6\%} \\
				single tower after fusion) & & \\ 
				Ours (S:VGG-16, T:VGG-16)  & 92.5\% & 65.4\% \\
				\hline
				
			\end{tabular}
		}
		\caption{Mean classification accuracy of best performing ConvNet approaches over three train/test splits on HMDB51 and UCF101. For our method we list the models used for the spatial (S) and temporal (T) stream.
		}
		\label{tab:SOTA}
		\vspace{-5pt}
		\resizebox {0.45\textwidth }{!}{ 
			\begin{tabular}{|l|c|c|}
				\multicolumn{3}{c}{} \\
				\hline
				IDT+higher dimensional FV \cite{Peng14} & 87.9\%  &	61.1\% \\  
				C3D+IDT \cite{C3DICCV2015} & 90.4\% & - \\ 
				TDD+IDT \cite{wang2015action} &  91.5\% & 65.9\% \\
				\hline
				Ours+IDT (S:VGG-16, T:VGG-M)&  92.5\% & 67.3\% \\
				Ours+IDT (S:VGG-16, T:VGG-16)& 93.5\% & 69.2\% \\
				\hline
			\end{tabular}
		}
		\caption{Mean classification accuracy on HMDB51 and UCF101 for approaches that use IDT features \cite{wangICCV13}.
			\label{tab:SOTA_IDT}
		}
	\end{center}
	\vspace{-20pt}

\end{table}

	\section{Conclusion}

We have proposed a new spatiotemporal architecture for two stream
networks with a novel convolutional fusion layer between the networks,
and a novel temporal fusion layer (incorporating 3D convolutions and
pooling).  The new architecture does not increase the number of
parameters significantly over previous methods, yet exceeds the state
of the art on two standard benchmark datasets.  Our results suggest the importance of learning correspondences between highly abstract ConvNet features both spatially and temporally. 
One intriguing finding is that there is still such an improvement by
combining ConvNet predictions with FV-encoded IDT
features. We suspect that this difference may vanish in time given far
more training data, but otherwise it certainly indicates where future
research should attend.

Finally, we return to the point that current datasets are either too small or too 
noisy. For this reason, some of the conclusions in this paper should be treated
with caution.


\vspace*{-3mm}
\paragraph{Acknowledgments.}
We are grateful for discussions with Karen Simonyan.
Christoph Feichtenhofer is a recipient of a DOC Fellowship of the
Austrian Academy of Sciences.
This work was supported by the Austrian Science Fund (FWF)
under project P27076, and also by EPSRC Programme Grant Seebibyte 
EP/M013774/1. The GPUs used for this research were donated by NVIDIA.

{\small
\bibliographystyle{ieee}
\bibliography{shortstrings,vgg_local,vgg_other,deep_actions}

}
\end{document}